\documentclass[sigconf]{acmart}

\AtBeginDocument{%
  \providecommand\BibTeX{{%
    \normalfont B\kern-0.5em{\scshape i\kern-0.25em b}\kern-0.8em\TeX}}}

\setcopyright{acmcopyright}
\copyrightyear{2019}
\acmYear{2019}

\begin{document}

\title{SwarmCloak: Landing of a Swarm of Nano-Quadrotors on Human Arms}

\author{Evgeny Tsykunov}
\email{Evgeny.Tsykunov@skoltech.ru}
\affiliation{%
  \institution{Skoltech, Moscow} 
}

\author{Ruslan Agishev}
\email{Ruslan.Agishev@skoltech.ru}
\affiliation{%
  \institution{Skoltech, Moscow}
}

\author{Roman Ibrahimov}
\email{Roman.Ibrahimov@skoltech.ru}
\affiliation{%
  \institution{Skoltech, Moscow}
}

\author{Luiza Labazanova}
\email{Luiza.Labazanova@skoltech.ru}
\affiliation{%
  \institution{Skoltech, Moscow}
}

\author{Taha Moriyama}
\email{moriyama@kaji-lab.jp}
\affiliation{%
  \institution{UEC, Tokyo}
}

\author{Hiroyuki Kajimoto}
\email{kajimoto@kaji-lab.jp}
\affiliation{%
  \institution{UEC, Tokyo} 
}

\author{Dzmitry Tsetserukou}
\email{D.Tsetserukou@skoltech.ru}
\affiliation{%
  \institution{Skoltech, Moscow}
}

\settopmatter{authorsperrow=4}

\renewcommand{\shortauthors}{Evgeny Tsykunov, et al.}

\begin{abstract}
We propose a novel system SwarmCloak for landing of a fleet of four flying robots on the human arms using light-sensitive landing pads with vibrotactile feedback. We developed two types of wearable tactile displays with vibromotors which are activated by the light emitted from the LED array at the bottom of quadcopters. In a user study, participants were asked to adjust the position of the arms to land up to two drones, having only visual feedback, only tactile feedback or visual-tactile feedback. The experiment revealed that when the number of drones increases, tactile feedback plays a more important role in accurate landing and operator's convenience. An important finding is that the best landing performance is achieved with the combination of tactile and visual feedback. The proposed technology could have a strong impact on the human-swarm interaction, providing a new level of intuitiveness and engagement into the swarm deployment just right from the skin surface.
\end{abstract}





\begin{teaserfigure}
    \includegraphics[width=\textwidth]{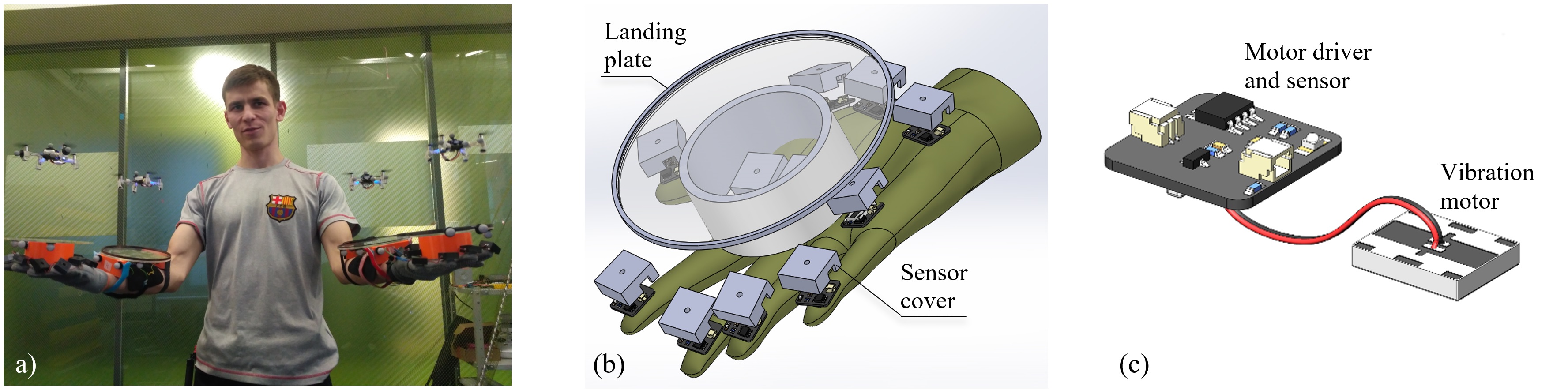}
    \caption{(a) Human operator lands four drones using landing pads, (b) hand-based landing pad, (c) sensor-vibrator unit.}
    \label{teaser_fig}
\end{teaserfigure}

\maketitle

\section{Introduction}

The swarm of drones is the subject of intensive interest of e-commerce companies and the scientific community. Fleets can contain thousands of robots and can be operated in both well prepared indoor and an occasional outdoor environment \cite{berg}.

For the human operator, it is often easier and faster to catch a small size quadrotor right in the midair instead of landing it on a surface in autonomous mode. There might be several reasons for it. For the outdoor applications, the landing surface is usually uneven and has a lot of dust, which could lead to a crash of the robots. Even when the landing spots (helipads) are well prepared, autonomous landing is not always the best solution due to position estimation errors, low robustness or high cost of a positioning system.

On the other side, the human body is widely used to control robot formation with gestures 
or hand motions \cite{Labazanova_2019,Tsykunov_2019}. 
Along with that, \cite{Scheggi_2014} demonstrated that tactile feedback could be effectively applied for the control and interaction with the swarm of robots. However, interaction strategies for formation takeoff/landing operations have not yet been considered, especially when the swarm of drones is landing on the human body.

To achieve interactivity during the swarm deployment, we developed the tactile interface that is used to deliver the information about the position of the drones relevant to the landing pads which are located on the human arms (see Figure 1). The location of the tactile stimulus reveals the position of the drone in the horizontal plane and stimulus intensity shows the distance to the robot in a vertical direction. Small drones are moving very fast, which requires high-speed communication channels. To reduce the latency, we propose to use the light emitted from the bottom of the drone to drive vibromotors.

\section{Principle and Technologies}
The single sensor-tactor unit (STU), shown in Figure 1(c), is based on HALUX technology \cite{Uematsu_2016} and it comprises a linear resonant actuator (LRA) (LD14- 002 by Nidec Copal Corporation), a phototransistor (PT19-21C by Everlight Electronics CO., Ltd.), and an oscillation circuit for LRA. LRA was selected for its fast response (less than 20ms). The resonance frequency of the oscillation circuit with LRA is 150Hz. Therefore, the vibration frequency is set to 150Hz. The amplitude of vibrations is modulated by the phototransistor. 

The prototypes of two types of devices (for hand and forearm) are shown in Figure 1 and Figure 2. The electronic circuit of each sensor-tactor unit is placed in the plastic cover which has a hole above photo-transistor for the light penetration (see Figure 1(b)). The plastic cover is used to protect sensor from surrounding undesirable light sources. The phototransistors are pointed upwards to detect the light emitted from the array of LEDs at the bottom of drone.

For the hand-based device shown in Figure 1(b), we placed all sensor-vibromotor units directly on the palm and finger pads. For the forearm-worn device (Figure 2), we fixedly attached all sensors to the ventral side of landing pads to robustly react on the light source, while the LRAs are placed directly on the forearm skin to generate strong tactile stimuli. The tactile pads are attached to human hand by Velcro tape.

The overall system consists of four landing pads and four Crazyflie 2.0 drones with integrated LEDs. In addition to the safety glasses, small size (9 cm$^2$) and weight (27 grams) of the quadrotor provide safety, which is required for applications that involve interaction with human. To track the quadrotors, we set up a Vicon motion capture system with 4 infrared cameras. We used the Robot Operating System (ROS) Kinetic framework to run the custom software.

During the experimental studies, SwarmCloak demonstrated several significant advantages over pure visual feedback. It was shown that the tactile feedback allows increasing accuracy of the landing pad positioning. It was also demonstrated that during the landing of two drones, tactile-visual feedback helped to considerably reduce the motion dynamics of the human head. Two-way ANOVA of drone positions showed a statistically significant difference for different feedback conditions. The number of drones does not significantly affect the performance of tactile feedback, in contrast to visual feedback. The best landing positions were achieved with the combination of visual and tactile feedback.

\begin{figure}[h]
  \centering
  \includegraphics[width=\linewidth]{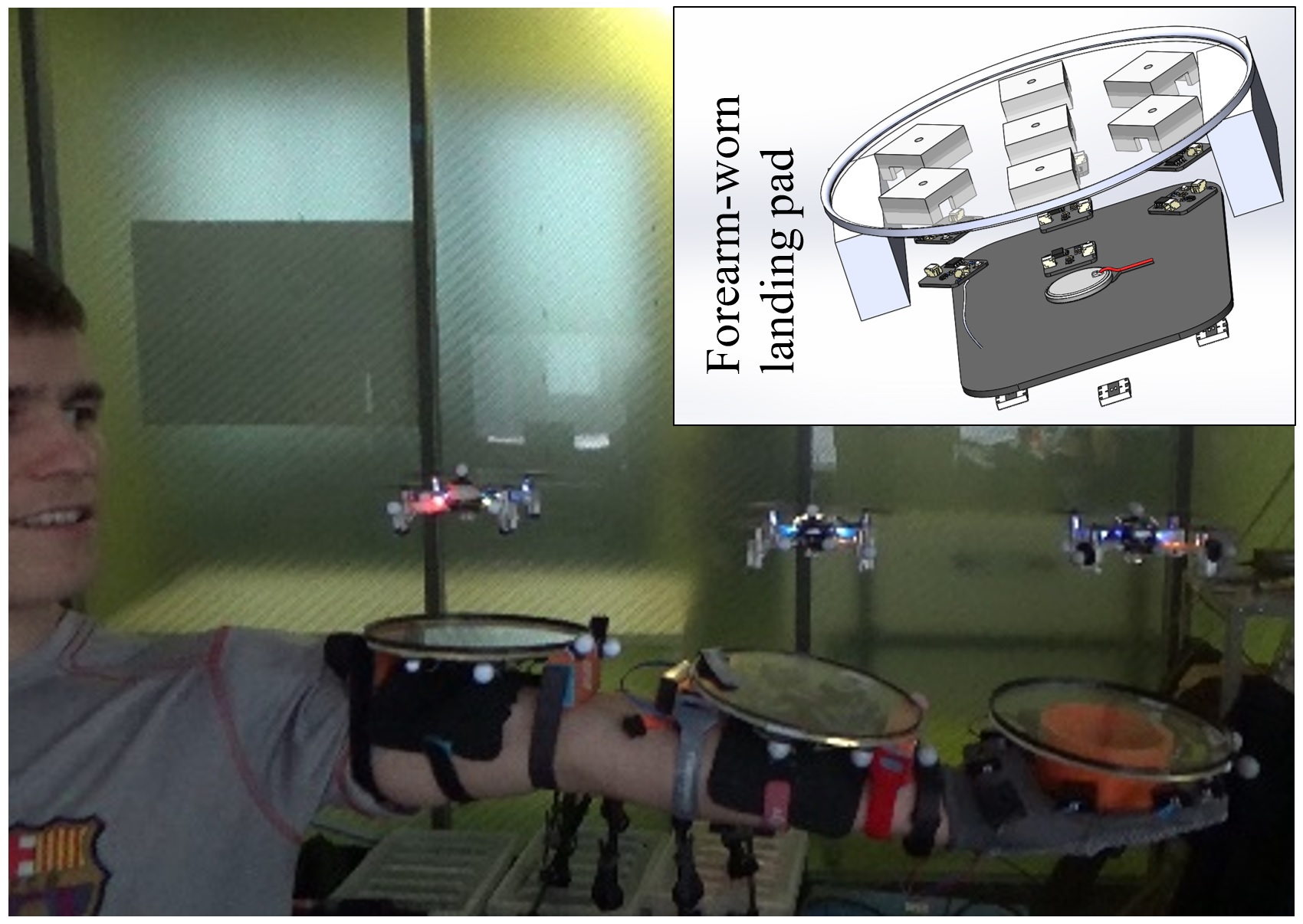}
  \caption{Landing of three drones on an arm.}
  \label{three_drones}
\end{figure}

\section{Applications and Future Work}

We proposed a novel system, in which the swarm is deployed in outdoor or indoor environments without a need for expensive and complex motion capture system. 

During the demonstrations, the users will wear two interactive landing pads with protective safety glasses and experience the landing of the fleet of drones on arms with tactile feedback.

Apart from the standalone landing of quadrotors, the technology might be used in various applications. The proposed device can significantly augment the perception of flying objects in Virtual Reality (VR) applications. Tactile sensations, such as bird landing or taking off from the human hands, can be simulated with SwarmCloak. Moreover, interaction with real or fictitious bioluminescence creatures, such as jellyfish or wood sprites of Tree of Souls from Avatar movie, can be implemented. 

A unique telecommunication system can be developed based on SwarmCloak technology. The partners can communicate through the distance by their avatars represented in VR and augmented by the swarm of drones. In this case, the swarm, which is capable of tactile interaction with a user in VR, might represent the skeleton structure of the human body flying in the air. This may bring a new level of immersion of VR communication and teleconferencing.

\bibliographystyle{ACM-Reference-Format}
\bibliography{sample-base}










\end{document}